\documentclass{article}
\usepackage{ijcai22}

\usepackage[ruled,vlined]{algorithm2e}
\usepackage{times}
\usepackage{graphicx}
\usepackage{mathrsfs}
\usepackage{amsfonts}
\usepackage{verbatim}
\usepackage{soul}
\usepackage{url}
\usepackage{amsmath}
\DeclareMathOperator*{\argmax}{arg\,max}

\usepackage[hidelinks]{hyperref}
\usepackage[utf8]{inputenc}
\usepackage[small]{caption}
\usepackage{graphicx}
\usepackage{listings}
\usepackage{amsmath}
\usepackage{amsthm}
\usepackage{booktabs}
\usepackage{color}
\newcommand{\Perf}{\textit{Perf}}

\title{Moses: Efficient Exploitation of Cross-device Transferable Features for Tensor Program Optimization}

\begin{document}

\author{
 Zhihe Zhao$^\dagger$, Xian Shuai$^\dagger$, Yang Bai$^\dagger$, Neiwen Ling$^\dagger$, Nan Guan$^\ddagger$, Zhenyu Yan$^\dagger$, Guoliang Xing$^{\dagger,\S}$
 \affiliations
 $^{\dagger}$The Chinese University of Hong Kong\\
 $^{\ddagger}$City University of Hong Kong\\
 $^{\S}$Corresponding Author\\
\emails
}
\maketitle

\begin{abstract}
Achieving efficient execution of machine learning models has attracted significant attention recently. To generate tensor programs efficiently, a key component of DNN compilers is the cost model that can predict the performance of each configuration on specific devices. However, due to the rapid emergence of hardware platforms, it is increasingly labor-intensive to train domain-specific predictors for every new platform. Besides, current design of cost models cannot provide transferable features between different hardware accelerators efficiently and effectively. In this paper, we propose Moses, a simple and efficient design based on the lottery ticket hypothesis, which fully takes advantage of the  features transferable to the target device via domain adaptation. Compared with state-of-the-art approaches, Moses achieves up to 1.53X efficiency gain in the search stage and 1.41X inference speedup on challenging DNN benchmarks.
\end{abstract}

\section{Introduction}

Efficient inference of deep neural networks (DNN) is of great importance for real-time AI applications such as autonomous driving and augmented reality, especially given that they usually run on embedded devices with limited compute power \cite{edgeml}.  Existing approaches usually rely on hand-optimized acceleration libraries, e.g., NVIDIA cuDNN and Intel MKL. However, they are vendor-specific, which cannot support a wide range of diverse hardware devices and require significant engineering efforts for tuning. 

\begin{figure}[!t]
\centering
\includegraphics[width=0.48\textwidth]{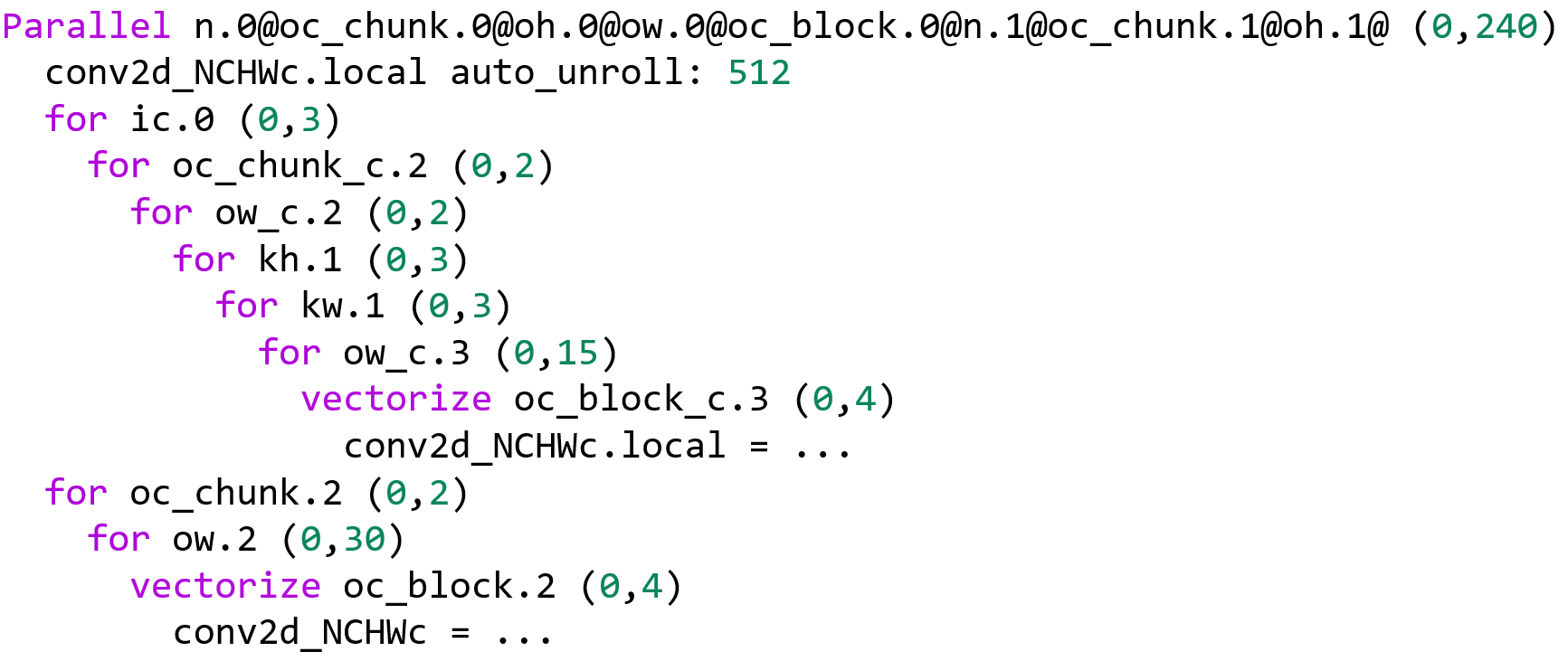}
\caption{An example of the program for a 2D convolution operator:
$Conv2d(in\_channels=3, out\_channels=64, kernel\_size=3, stride=1, padding=0, bias=False)$ produced by TVM.}
\label{fig:prog}
\end{figure}

Typically, the tensor computation inside deep learning operators is implemented by a set of compute-intensive nested loops. 
For example, Figure \ref{fig:prog} shows an optimized tensor program of a 2D convolutional operator, which consists of multiple for-loops and involves three schedule primitives: blocking, unrolling, and vectorization. However, generating such high-performance tensor programs from a given high-level expression is extremely difficult, as the optimal organization and the parameters of the for-loops can vary significantly for different devices. Therefore, to accelerate the end-to-end model inference on various hardware platforms, existing DNN compilers first generate a large space of configurations, and then search for the best-performing one based on on-device measurements \cite{autoTVM}. This process is termed as auto-tuning. 

Although DNN compilers can produce optimized programs for DNN models on specific platforms, they suffer excessively long search time. For example, although AutoTVM can outperform nearly 2× over default TensorFlow 
on ResNet-18, the auto-tuning time can take up to tens of hours on embedded devices (e.g. NVIDIA Jetson TX2).
To reduce the time-consuming on-device measurements, the tensor program optimizer employs a cost model to directly predict the performance of the potential candidates in the search space. However, training a cost model offline still requires a large number of measurements. For instance, Tenset \cite{tenset} provides a tensor program performance dataset collected from 6 devices, containing 52 million program performance records. Based on this dataset, it was shown that the cost model can be transferred between two Intel CPUs by fine-tuning. As a result, the online search time can be reduced without sacrificing the quality of optimized tensor programs.
However, when two hardware platforms differ significantly in architecture, such a vanilla fine-tuning approach would 
fail to learn the runtime behaviors of a new device, and hence performs poorly at generating high-performance tensor programs, as evidenced by our results in Section 4.  

Previous efforts trying to address this challenge mainly focus on either designing a new cost model \cite{costmodel2}\cite{costmodel_tpu}, or exploring effective search algorithms during auto-tuning \cite{CHAMELEON}\cite{adatune}. However, these approaches still need large amounts of iterations during the search and the performance is usually highly program-dependent. 
Motivated by these challenges, we propose Moses, a novel cost model adaptation framework based on the lottery ticket hypothesis \cite{lottery}, which can adapt the trained cost model from a source device to a new target device with high efficiency. We summarize the contributions of this work as follows:

\begin{itemize}
    \item [1)] To the best of our knowledge, Moses is the first work that achieve highly efficient for auto-tuning between different hardware platforms with transfer learning.  
    Moses enables the DNN compiler to generate optimized tensor programs with significantly shorter search time for a new device. 
    \item [2)] We propose a novel approach that can automatically identify the transferable hardware-independent parameters of a pre-trained cost model, and achieve cross-device cost model adaption via fine-tuning. 
    \item [3)] We conduct comprehensive experiments and show that Moses is a general and effective approach for diverse hardware platforms and DNNs.  
\end{itemize}

\section{Related Work and Background}
\subsection{DNN Compilers.}
General DNN compilers optimize the computation flow of DNN tasks in two levels: graph-level and tensor-level. Some notable compilers are TVM \cite{TVM}, TASO \cite{TASO}, XLA \cite{xla} and Halide \cite{hilide}. 
These compilers either utilize compiler techniques such as graph substitutions to optimize the intermediate representation (IR) level graph \cite{autogtco}, or focus on tensor program optimization using heuristic and learning-based algorithms to joint-optimize the polyhedral patterns for DNN. 
Building on these DNN compliers, recent works treat the optimization process as a black box and propose some advanced searching or cost model training approaches based on run-time information \cite{autoTVM}\cite{CHAMELEON}\cite{adatune}\cite{protuner}.

\vspace{-1ex}\subsection{Auto-Tuning with Cost Models}
\begin{figure}[!t]
\centering
\includegraphics[width=0.47\textwidth]{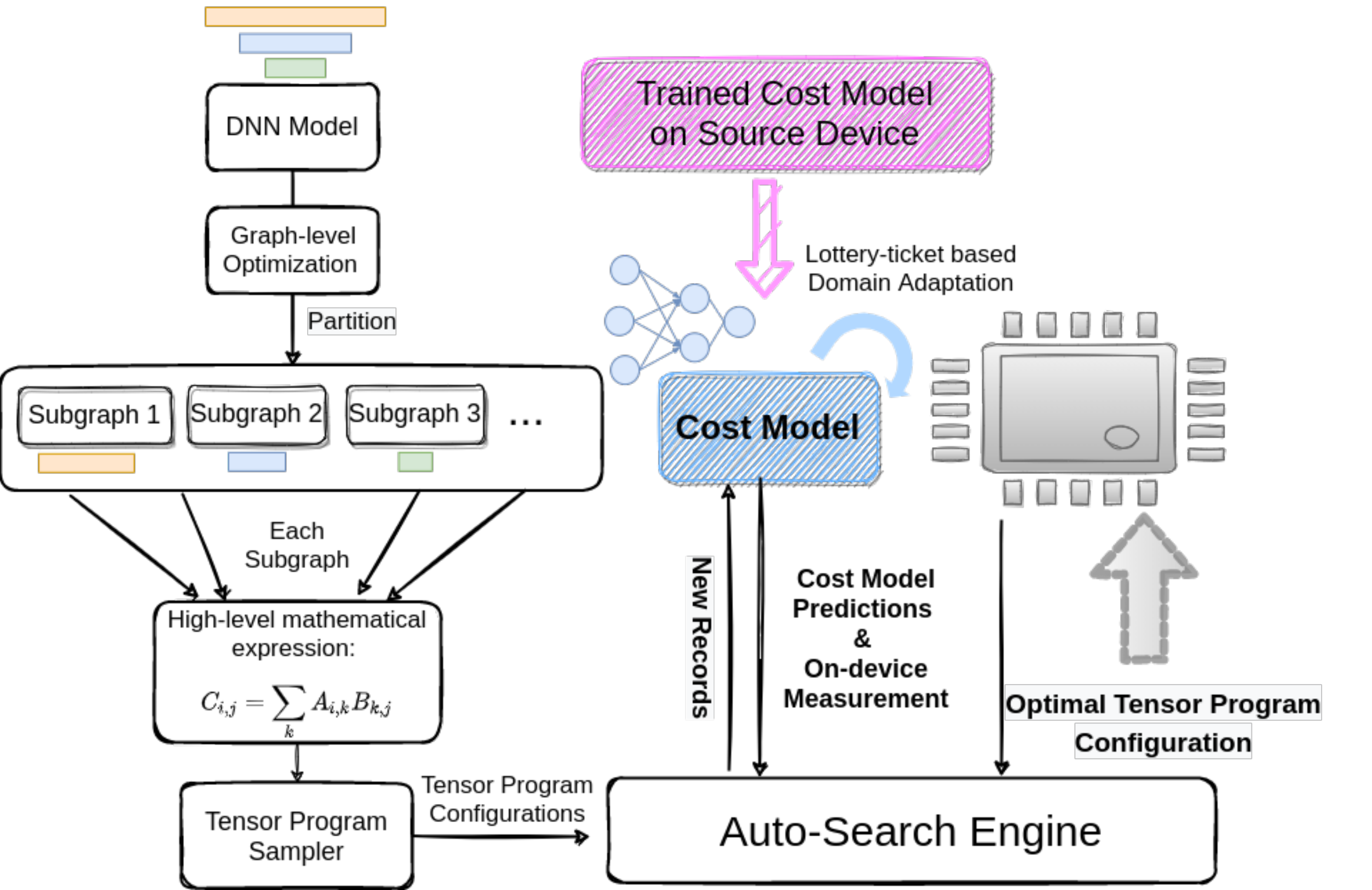}
\caption{The complete pipeline of automatic tensor program generation for a given neural network.}
\label{fig:overview}
\end{figure}
Figure \ref{fig:overview} shows the pipeline of a search-based low-level tensor code generation used by TVM \cite{TVM}. The intact neural network is first partitioned by the graph-level optimizer. For instance, the ResNet-50 will be be divided into 29 subgraphs and each subgraph normally includes one or two convolutional layers. Given an intact mathematical expression or a computational graph in a high-level mathematical expression, the search algorithm will search for the best low-level tensor implementation for the target hardware. 
Usually, the search space is in the order of millions for CPUs and billions for GPUs, as there are a variety of schedule primitives such as tiling, unrolling, vectorization, parallelization, and thread binding. Each schedule primitive can involve multiple tunable knobs. 
On one hand, such a large search space enables the automatic tensor compiler to find the program that is better than the hand-optimized implementation. On the other hand, the large search space can incur significant search time, especially for embedded devices with limited computation power. 

To accelerate the searching process, TVM introduces the cost model to directly predict the time cost of the innermost non-loop program rather than extensively measuring program's runtime. 
In this paper, we adopt the 164-$d$ features in Ansor \cite{ansor} to depict the program. In each iteration, based on predictions from the cost model, a batch of candidate programs are sampled by an evolutionary search engine. Then, TVM measures the actual time costs of these sampled programs. Finally, the measurements are added to the training data of the cost model. Therefore, the cost model and searched tensor program are improved iteratively.

Notably, TVM includes two search frameworks: AutoTVM \cite{autoTVM} and Ansor \cite{ansor}. AutoTVM requires hand-written scheduling templates for searching, while Ansor is a fully automated framework. Due to the large number of hardware platforms and the substantial efforts to develop templates, we use Ansor in this paper.


\vspace{-1ex}\subsection{Cross-Device Cost Model Adaptation}
As discussed in the previous section, a major practical drawback of the current automatic tensor program optimization approach \cite{autoTVM} is the extremely long search time. Recent work such as \cite{CHAMELEON} provides a breakdown of the search time and shows that the time for on-device measurements dominates. 
Therefore, a solution to shorten the online searching time is to collect an comprehensive tensor program performance dataset offline, and pre-train a cost model that can be directly utilized \cite{tenset}. 
However, the cost model is device-specific, as its input features include configuration knobs that are closely related to the hardware architecture such as the lengths of BlockIdx and ThreadIdx. When two architectures are significantly different (e.g., server GPUs and mobile GPUs), the traditional transfer learning approach may fail.


\begin{figure}[!t]
\centering
\includegraphics[width=0.47\textwidth]{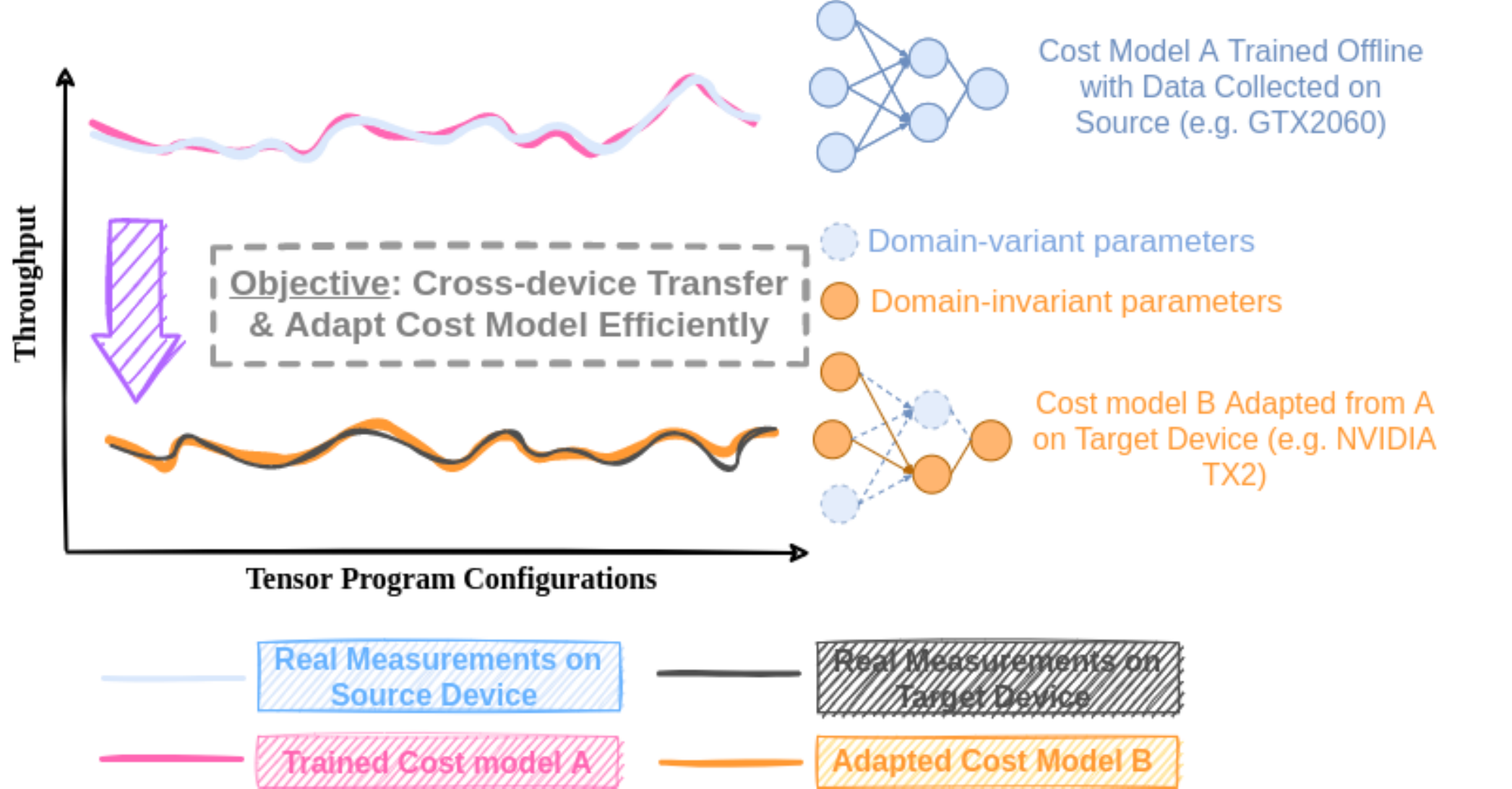}
\caption{
Given a cost model A trained on the source device, we aim to obtain a cost model B that can accurately predict the throughput of the target device under various tensor program configurations. }
\label{fig:3}
\end{figure}



\section{Moses}
In this section, we present the overview, problem formulation and design of Moses.

\subsection{Overview}
We propose Moses, a novel cross-device cost model transfer approach based on the lottery ticket hypothesis \cite{lottery}. As shown in Figure \ref{fig:3}, we transfer the cost model trained on source devices (e.g., Server GPUs) to target devices (e.g., mobile GPUs) by only fine-tuning portion of the model parameters while keeping the rest of parameters deactivated. The rationale of our design is two-fold. First, to accelerate the online searching instead of collecting a dataset for every new device offline, we need to take advantage of the cost model pre-trained on source devices rather than training a new model from scratch. Second, as vanilla fine-tuning approaches may fail due to substantial architecture changes, we have to utilize the prior knowledge from the pre-trained cost model wisely. Specifically, we leverage the lottery ticket hypothesis to identify the transferable parameters in the cost model space \cite{lottery_para}. By distilling the transferable parameters and dropping out the untransferable ones, we can not only shrink the size of the cost model but also reduce the device-specific information, which accelerates the training of the cost model and improves the generalizability. 

\vspace{-1ex}\subsection{Problem Formulation}
The Auto-tuning process in a DNN compiler aims to generate a large search space of tensor program configurations and to find the optimal one based on the on-device performance measurement records \cite{metatune}. We denote the transformation pass of tensor programs by $t$ and the cost model by \textit{C} . During the search process in auto-tuning, the compiler picks the top-k (where $k$ can be one) programs with the performance predictions from the cost model for each task. The process can utilize mixed on-device measurements and cost model predictions. If the on-device measurement cost is large, the process can completely rely on the cost model; Function $\textit{Perf()}$ represents the run-time hardware measurement (throughput, GFLOPs); Function $\textit{C()}$ represents the cost model predictions on codes performance; \textit{i} is an input task, which can be a computing subgraph in the input DNN model. For example, SqueezeNet consists of 23 tasks (a.k.a, the subgraphs) in total. Note that a subgraph is a unit with the finest granularity during the compilation process; \textit{$\Psi$} is the set of possible tensor program configurations of a task which consists of a combination of parameters called \emph{knobs}; $g()$ represents the tensor programs generating functions with knobs and the transformation pass as inputs. The approximation of a cost model function to real hardware measurement can be denoted by $C()\sim \Perf()$; Thus, given an input subgraph \textit{i}, the objective of the auto-tuning process is finding the optimal combination of knobs $\psi^{*}$ to maximize the performance defined as below:
\[\psi^{*} = \argmax_{\psi \in \Psi} \Perf(g(\psi,t))\eqno{(1)}\]
Precise estimate of the performance of tensor program candidates can effectively reduce the time-consuming on-device measurements. Thus, the objective of the cost model is to minimize the difference between the predictions and the real-world measurements, which is:
\[\min \left \| C(g(\Psi,t)|\psi,\theta) - \Perf(g(\Psi,t))\right \|\eqno{(2)}\]
Here, $\theta$ represents the parameter sets in the trained cost model. We train a high-performance cost model is to find the best parameter configuration $\Theta^{*}$, which can efficiently guide the auto-tuning search process.\\
The on-device measurements especially for embedded/mobile devices are extremely time-consuming. For example, the time consumption for on-device measurements of a VGG16 model can be up to 10 hours on an NVIDIA TX2. In Section 3.3, we design a cross-device domain adaption method, which transfers the cost model trained on the source device to the target device. The objective of the cross-device domain adaptation is to find a new parameter set $\Theta^{\dagger}$ 
which can minimize the difference between the real-world measurements and cost model predictions, and thus to guide the tuning process for each subgraph on the target device in an adaptive manner. 

\vspace{-1ex}\subsection{Feature Space Representations}
In our problem, the feature space is independent of hardware architecture while the outputs of the cost model (throughput predictions for different tensor program configurations) are dependent, shown as:
\[H\left\{ \mathscr{X} \right\}\equiv H\left\{ \mathscr{X}_{DIV} \right\} + H\left\{ \mathscr{X}_{DV} \right\}\eqno{(3)}\]
where $H{}$ maps the hidden feature space representations, $\mathscr{X}_{DIV}$ and $\mathscr{X}_{DV}$ present the decoupled feature space: hardware-independent and hardware-dependent information, respectively. 
The SOTA \textit{Tenset} \cite{tenset} shows that the model fine-tuning on the target domain (devices) is effective only when the architecture difference between two devices is small. However, Tenset does not provide any specific solutions, e.g., the cost model domain adaptation between a server level GPU and a mobile level GPU. In order to solve this problem, we propose to leverage the following lottery ticket hypothesis in this paper: only part of the parameters in the trained cost model on source devices are essential for learning the hardware-independent knowledge. In other words, Only part of the information needs to be adapted to the target device while other parameters tend to fit the domain. Motivated by this, we can transform our problem into the following question: how to learn domain invariant parameters to minimize the domain discrepancy brought by the hardware difference?\\ 

\vspace{-1ex}\subsection{Lottery-Ticket-Based Cross-Device Adaptation}
The training data for cost model updating is collected online during the search, which makes the search very time-consuming due to inevitable on-device measurements. We denote the set of labeled tensor program records (program knobs, throughput) on source device by $S = {\{(x_{s}^{i}},{y_{s}^{i})\}}^{m}$, where $m$ is the training data points on source device and $(x_{s}^{i}$, $y_{s}^{i})$ is the $i_{th}$ record and its corresponding label respectively in domain $s$. Now given a target unlabeled program performance records set $T = \{{(x_{T}^{j})}\}^{k}$ with different configurations of program knobs $x$ sampled from target device $D_{T}$, in which a small set of $\widehat{T} = \{{(x_{\widehat{T}}^{j}},{y_{\widehat{T}}^{j})}\}^{n}$ can be collected by on-device measurements (which are conducted typically under a given time budgets due to the time-consuming nature). We have $n \ll k \ll m$. Formally, The expected domain adaptation error on target device can be defined as $\varepsilon(h_{Target}(\Theta^{\dagger}))$ where $h \in H$, a hypothesis that learned from $\widehat{T}$, thus the following inequity holds:
\[\varepsilon(h_{Target}(\Theta^{\dagger}))\le \varepsilon(h_{Source}(\Theta^{*}))\]
\[+dist(\mathscr{D}_{S}(\mathscr{X}),\mathscr{D}_{T}(\mathscr{X}))+\varphi, \eqno{(4)}\]
In the terminology of semi-supervised domain adaptation, $dist$ represents the distribution discrepancy over the cross-device domains, 
and $\varphi$ represents the ideal error or risks achieving cross-device domain adaptation. Since the feature representations are influenced by hardware difference in our problem, to achieve the cross-device domain adaptation, instead of minimizing the distance between feature representations and their resulting data distribution discrepancy, we show the effectiveness of bound minimization strategy on solving the cross device domain adaptation and propose to find the bound limitation by optimizing the labeling black-box functions.
Such an approach is inspired by the lottery ticket hypothesis, which was originally proposed in the context of model compression, showing that only part of parameters are fit for model generalization \cite{lottery}.

Similar as in \cite{lottery}, we show experimentally in this paper the same hypothesis holds in our problem (see Section 4). That is, there exists a super-subnet, named winning ticket, with a set of essential parameters of the trained cost model on source devices, which would be the domain invariant information. In other words, training from a super-subnet on the target device would achieve the optimal transfer performance in our cross-device domain adaptation problem. 

We name parameters in these super-subnets as transferable parameters, and the remaining set of parameters as untransferable parameters. A key question is how to distill these transferable parameters during each subgraph auto-tuning stage. We identify the distilling boundary criterion $\xi(ph)$ as below:
\[\xi(i) =\left| w(ph)*\nabla w(ph) \right| \eqno{(5)}\]
where we denote the tuning phase of the subgraphs by $ph$. $w(ph)\in W(ph)$ represents the parameter weights and its gradient is $\nabla w(ph)$. If $\xi(ph)$ is larger than a certain threshold $\vartheta$ (e.g., 0.5), the corresponding $w(ph)$ can be viewed as transferable parameters. In contrast, $w(ph)$ would be regarded as domain-variant parameters if $\xi(ph)$ is small(e.g., close to zero). We also provide a ranking mechanism here, specifically, we rank these parameters based on their cross-domain importance according to Eq.(5). Thus the users can set the transferable parameters ratio manually. We iteratively update the boundary of domain-invariant parameters as well as variant parameters and update these invariant ones during each online training epoch to learn invariant representations to achieve minimization of bound limitation $\varphi$. Lastly, we adopt the adversarial loss training objective function $L_{inv}(S,T)$ defined as below:
\[L_{inv}(S,T) = L_{x\sim p(S)}(lg(b(w(ph,inv))))\]
\[+\beta L_{x\sim p(T,\widetilde{T})}(lg(1-b(w(ph,inv))))\eqno{(6)}\]
where $b()$ represents the labeling black-box functions, $\beta$ as the coefficient that control the entropy effects (usually set as a small number). In such a way, the parameters with higher gradient flows, representing more beneficial to the domain-invariant information learning process, are considered. As for the parameters that represent domain variant information, We update theses parameters with a weight decay mechanism as penalty, which is defined as:
\[w_{v}(ph+1)\rightarrow w_{v}(ph)-\alpha(wd(w_{v}(ph)))\eqno{(7)}\]
where $\alpha$ is the learning rate, $w_{v}(ph)$ represents the domain-variant parameters and function $wd()$ represents the weight decay process for each updating phase. 

\vspace{-1ex}\subsection{Adaptive Tuning Data Partition}
To gather on-device measurements efficiently and maintain the performance of the online domain adaptation cost model, we use an adaptive controller (AC) module to early terminate the hardware tuning data collection stage. The basic idea of AC is to statistically analyze the certainty of online training cost model. For a give subgraph $s$ which is to be tuned, we initially divide the total tuning tasks into \emph{tuning tasks for online training $t_{train}$} with hardware measurement data collection and \emph{tuning tasks for cost model predictions $t_{pred}$} with a ratio of $p$. We further divide $t_{train}
$ into $q \in {1,2,..,q}$ batches and collect both on-device measurement records $C(t_{train}(s))$ and $\Perf(t_{train}(s))$. Then, we use the coefficient of variations (the standard deviation divided by the mean) formulated as $CV = \frac{\sigma(C(t_{train}(s))_{1},C(t_{train}(s))_{2}..C(t_{train}(s))_{q})}{\mu(C(t_{train}(s))_{1},C(t_{train}(s))_{2}..C(t_{train}(s))_{q})}$ to dynamically estimate the certainty of the existing cost model, and terminate the hardware measurement phase in advance if the value is smaller than a certain value. We empirically set these hyper-parameters based on multiple trials in our experiments.

\vspace{-1ex}\subsection{Putting Everything Together}
In previous sections, we described the design details of the lottery-ticket based cross-device cost model transfer and the adaptive online training data partition mechanism. We now put these components together and summarize the working flow of Moses. \\
\textbf{Step 1. Pre-training a cost model on the source device:} We pre-train a cost model offline using the dataset of on-device measurement records from the source device. This dataset includes randomly generated tensor programs for widely deep learning models. \\
\textbf{Step 2. Transferring the trained model to the target device:} The learned cost model from source device is directly transferred to the target device in this step, to guide the search stage during auto-tuning.\\
\textbf{Step 3. Adaptive training data partition with the AC module:} For each tuning task, we dynamically control the hardware measurement costs by using the AC module, with which the portion of on-device measurements can be adjusted by the evaluations of cost model performance in that epoch.\\
\textbf{Step 4. Online updating the cost model with iterative pruning:} 
For each tuning task, 
we divide the parameters of the cost model into domain-invariant ones and domain-variant ones based on the calculation of $\xi(i)$, and update the domain-invariant parameters with gradient decent while letting the rest gradually decrease to zero due to weight decay. \\ 
During the auto-tuning process, Moses keeps updating the cost model in an adaptive and iteratively manner based on the collected hardware measurements records, while the search algorithms keep querying the newest cost model for efficient explorations of optimal program configurations.

\section{Experiments}

\begin{table*}[t]
    \caption{A summary of comparisons of CMAT under small and large trials. S, R, M and B refer to four DNNs mentioned in Fig. \ref{fig:search}. }
    \centering
    \scalebox{1}{
    \begin{tabular}{c|c|c|c|c|c|c|c}
    \textbf{CMAT (\%)} & \textbf{2060-S} & \textbf{2060-R} & \textbf{2060-M} & \textbf{2060-B} & \textbf{TX2-S} & \textbf{TX2-R} & \textbf{TX2-M}\\  \midrule[1pt]
    Small Trials (200) & 57.2  & 19.6  & 105  & 66.7  & 28.7  & 66.4  & 64.5 \\ 
    Large Trials (20000/5000)  & 48.1  & 32.7  & 45.8  & 87.4  & 44.7  & 53.1  & 45.9\\ \hline
    \end{tabular}
    }
    \label{dataset_table}
    
\end{table*} 
In previous sections, we describe how we enable the cross-device cost model domain adaptation and make the auto-tuning process on a new device more adaptive 
In this section, we evaluate the effectiveness of Moses, with our proposed lottery-ticket based cost model adaptation method.
We implement Moses as a plug-in cross-device cost model adaptation tool in TVM auto-tuning \cite{TVM}. Specifically, the cost model fine-tuning is integrated with the tensor programs random sampling and an evolutionary search algorithm \cite{ansor}. The training of the cost model is implemented in PyTorch. We set the max epoch to 30. We set the initial learning rate $alpha$ to 0.001, the distilling boundary criterion threshold $\vartheta$ to 0.5.

\subsection{Generated Dataset for Embedded Devices}
As mentioned before, learning-based tensor compilers can greatly boost DNN model inference performance. At the core of the auto-tuning process in these compilers, is a cost model which estimates the performance of the combinations of tensor representation knobs on different devices with an input DNN model. To perform the evaluation and ease the training efforts on cost models especially on embedded devices. In this paper, we collect a comprehensive tensor program dataset for two embedded devices: NVIDIA Jetson TX2 and XAIVER. We collect tasks from over 50 DNN models including the popular mobile transformers (e.g. mobileViT \cite{mobilevit}). More than ten million of program records are included in this dataset. 

\subsection{Experimental Settings}

Our experiments are conducted on NVIDIA GeForce GTX 2080 and NVIDIA Jetson TX2 with Pascal GPU architecture with 256 NVIDIA CUDA cores. The current deployments such as \cite{ansor,flextensor,adatune} optimize the input DNN models in a subgraph basis, which means to optimize the fused operators one by one. Thus the the processes are independent of the DNN model, which means we do not need to include multiple complicated DNN models to validate Moses. The current popular operators of DNN models for both academia and industry can be summarised as: convolutional layers, depthewise separable convolutional layers; multi-head attention module; residual block and other types of layers such as fully connected layers and pooling layers. We include four DNN models in our experiments: ResNet-18, MobileNet, BERT-base and SqueezeNet. We use the default settings for other hyper-parameters provided by Ansor \cite{ansor}. As for the backbone of the cost model, we choose the representative one used in Ansor, which is an MLP with two hidden layers, with 512 neurons for each. We train the MLP cost model with ranking loss on NVIDIA Tesla K80, which is noted as the source device (domain), based on the dataset provided by Tenset \cite{tenset}. The two main domain adaptation tasks we validate are $K80 \rightarrow 2060$ and $K80\rightarrow TX2$. 

\begin{figure}[t]
\centering
\includegraphics[width=1\linewidth]{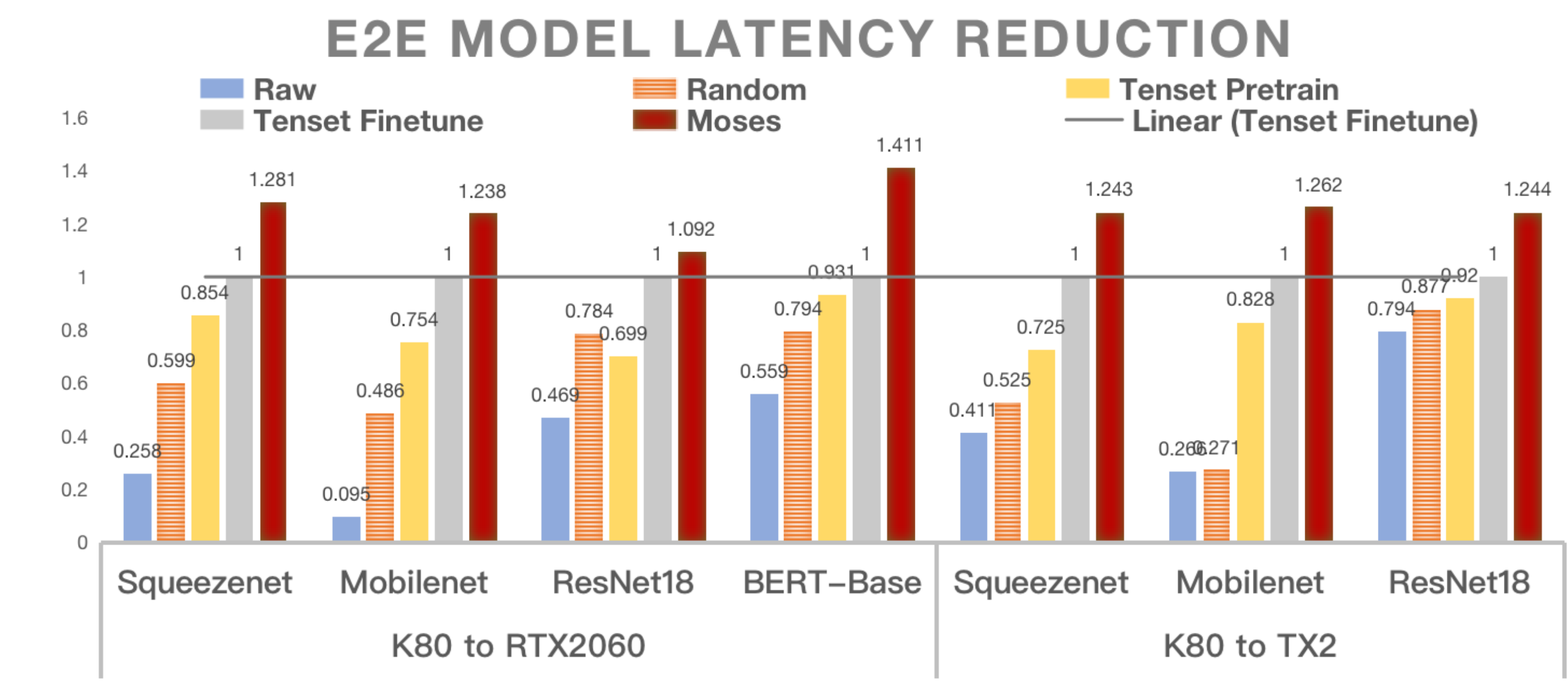}
\caption{End-to-end DNN model inference latency reductions GAIN comparisons among MobileNet, ResNet18, BERT-Base and SqueezeNet over two domain adaptation baselines.}
\label{fig:lat}
\end{figure}
\begin{figure}[t]
\centering
\includegraphics[width=0.96\linewidth]{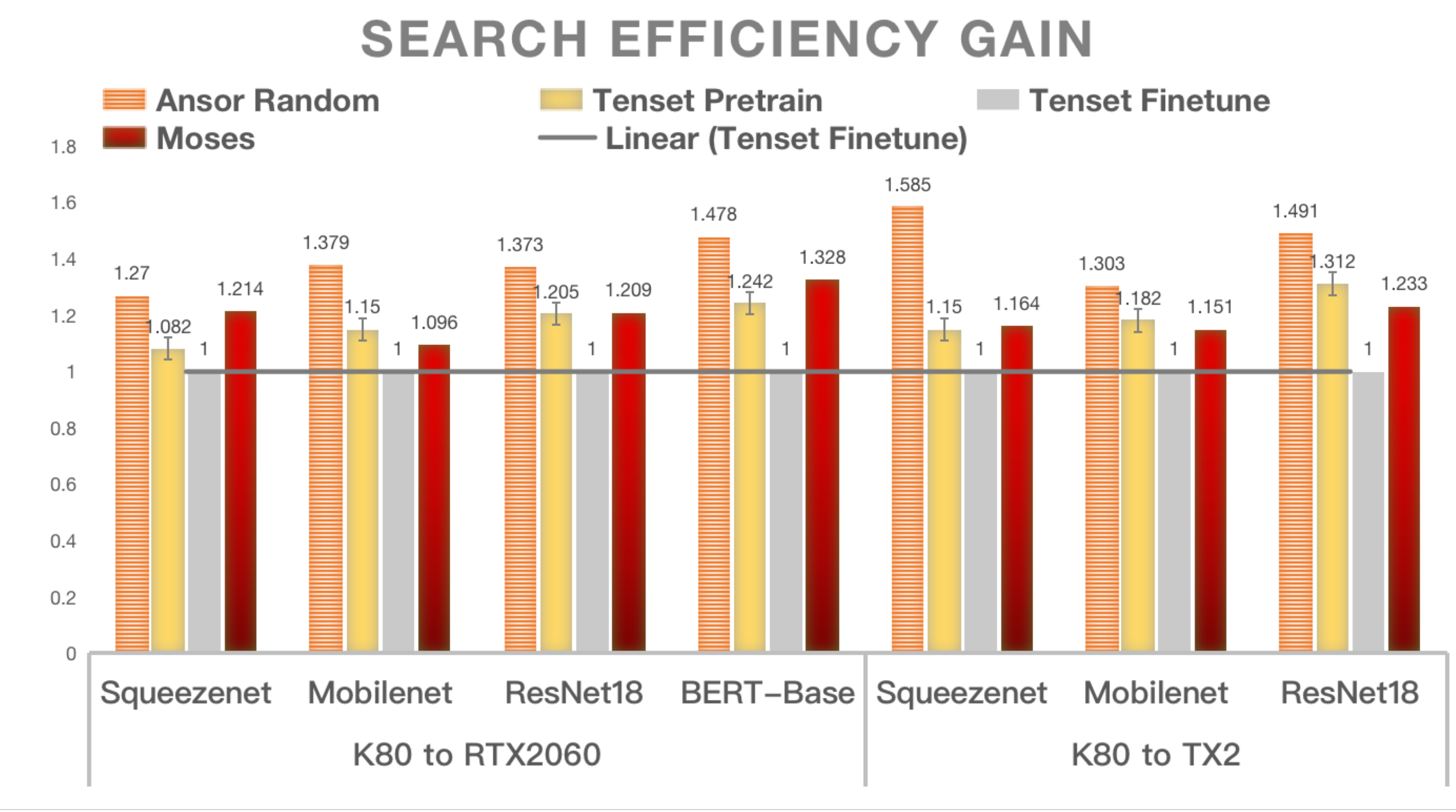}
\caption{Auto-tuning search efficiency GAIN comparisons among MobileNet, ResNet18, BERT-Base and SqueezeNet over two domain adaptation baselines.}
\label{fig:search}
\end{figure}

\vspace{-1ex}\subsection{Evaluation Metrics}
We use the end-to-end latency/throughput and the end-to-end search efficiency of auto-tuning  as two main metrics. Specifically, we measure the obtained speedups of tuned tensor programs and the reductions of searching time of an input DNN model over other baselines including the stat-of-art cost model transfer method provided in Tenset. We also introduce a concept named Cost Model \& Auto-tuning Efficiency Gain Score (CMAT) to evaluate the cost model influence on the end-to-end inference performance at the same time, defined as:
\textit{CMAT = (Gain on Search Efficiency * Reduction on Tuned Model Latency-1)*100\%}. As CMAT considers both search efficieny and end-to-end inference latency, it is an effective metric for evaluating the overall cross-device cost model adaptation performance.

\vspace{-1ex}\subsection{Results}
The main results we provided aim to answer the following questions: 1). Can Moses transfer the trained cost model to different hardware platforms and outperform other online fine-tuning method provided by Tenset? 2). Can Moses accelerate the auto-tuning process by adaptively scheduling the portions of on-device measuring records trials?
To answer these questiones, we compare Moses with four baselines:
\begin{itemize}
\setlength{\itemsep}{0pt}
\setlength{\parsep}{0pt}
\setlength{\parskip}{0pt}
\item [1)] Raw: inference with original CUDA acceleration.
\item [2)] Ansor-Random \cite{ansor}: randomly initialize the cost model and train it from scratch during the auto-tuning.
\item [3)] Tenset-Pretrain: pre-train a cost model on TenSet dataset and directly apply it to the target device without fine-tuning.
\item [4)] Tenset-Finetune: utilize the cost model pre-trained on TenSet dataset and then perform the vanilla online fine-tuning. 
\end{itemize}

\begin{figure}[t]\centering
  \includegraphics[width=1\linewidth]{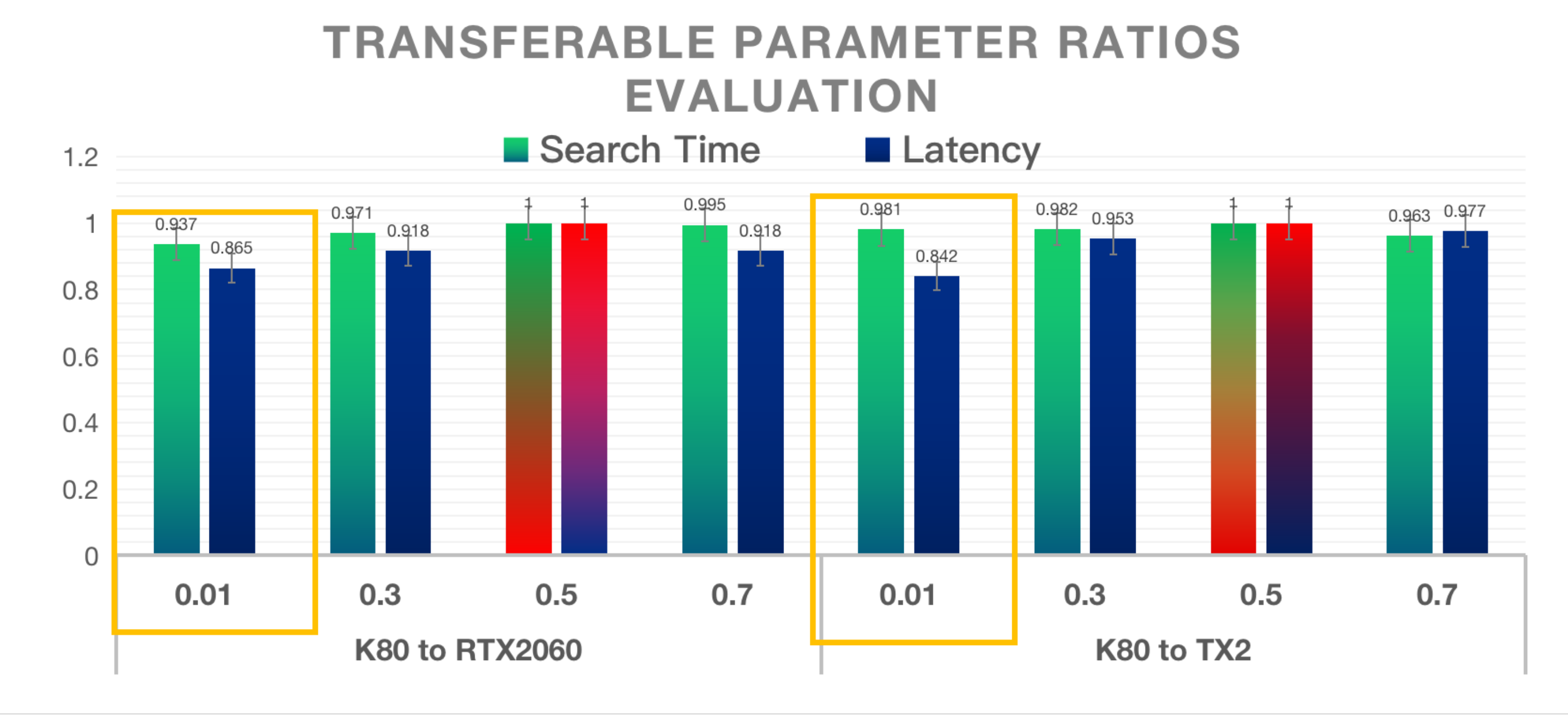}
  \caption{An illustration of Moses performance with a wider ratios of transferable parameters, the yellow box shows the results of ratio=0.01.}
  \label{para ratio} 
\end{figure}
\vspace{-1ex}\subsubsection{Inference Time \& Search Efficiency Comparisons.} Fig. \ref{fig:lat} shows the comparison of the final end-to-end inference time of the input DNN model optimized on each baseline. Moses achieves up to 41.1\% faster inference speed over Tenset-Finetune and up to 53\% higher speed over Tenset-Pretrain on the $K80\rightarrow 2060$ baseline. Moses also achieves up to 26.2\% over Tenset-Finetune and up to 52\% over Tenset-Pretrain on the $K80\rightarrow TX2$ baseline, respectively. Overall, Moses yields the best inference performance among all other configurations and algorithms. Fig. \ref{fig:search} shows the auto-tuning search efficiency gains comparisons over these baselines. Moses also outperforms all other baselines for both $K80 \rightarrow 2060$ and $K80 \rightarrow TX2$ settings. It can also be observed that, for some input DNN models such as SqueezeNet and MobileNet, Ansor-Random and Tenset-Pretrain could be more efficient than Moses. This is because these baselines provide no online learning during the auto-tuning. Therefore, the corresponding end-to-end model inference latency of these models can be greatly lower than Moses. The evaluation results show that, the search efficiency gain of the $K80 \rightarrow 2060$ setting can be up to 47.8\% while up to 58.5\% for the $K80  \rightarrow TX2$ setting. This is because the on-device data collection costs on TX2 is much higher than on RTX2060.

\vspace{-1ex}\subsubsection{CMAT Score Comparisons.} Table. \ref{dataset_table} shows the superior comprehensive performance of Moses over both small (200) and large (20000 for 2060, 5000 for TX2) number of trials across all input DNN models. As mentioned above,  although Tenset achieves 15\% auto-tuning efficiency gain on MobileNet based on the $K80\rightarrow 2060$ setting, which is better than Moses (9.6\%), the corresponding CMAT is -14.75\% over Tenset fine-tuning, which is much worse than Moses (up to 45.8\%). We can observe that for some cases (e.g. 2060-S), the CMAT gain under the small-trial setting can even be better than the large-trial one, due to the characteristics of the heuristic searching algorithm embedded in the auto-tuning component in TVM. The base performance of Tenset-Finetune can be extremely low with no prior knowledge during the transfer process of the cost model.

\vspace{-1ex}
\subsection{Ablation Study: Ratio of Transferable Parameters.}
Here we provide more analysis on the ratio of transferable parameters. Fig. \ref{para ratio} shows the results of Moses on a wider setting of transferable parameters ratio: $\{0.01, 0.3, 0.5, 0.7\}$. According to the end-to-end performance results, we can observe that the optimal performance can be around 0.5. Generally speaking, the $std$ value for settings of $\{0.3, 0.5, 0.7\}$ ratio is not large, which suggests that, the optimal performance produced by different ratios is not sensitive to the ratio setting when it is ranging from 0.3 to 0.7.

\section{Conclusion \& Future Work}
We present Moses, a new framework to optimize the auto-tuning process in DNN compiler. Specifically, our approach enables cross-device domain adaptation of a trained cost model by updating the domain invariant parameters during online learning, which greatly improves the efficiency of auto-tuning process and the end to end throughput of tuned tensor programs on the target device. 
Besides, we generate a large-scale program performance dataset on two embedded GPUs for learning based DNN compilers. Our future work includes extending Moses to support knowledge transfer from the cross-subgraph tensor optimization perspective. 

\bibliographystyle{named}
\bibliography{ijcai22}

\end{document}